\documentclass[runningheads]{llncs}

\usepackage[final,year=2026,ID=7827]{eccv}

\usepackage{eccvabbrv}

\usepackage{graphicx}
\usepackage{booktabs}

\usepackage[accsupp]{axessibility}  

\usepackage{hyperref}

\usepackage{orcidlink}
\usepackage{multirow}
\usepackage{colortbl}
\usepackage[table]{xcolor}
\usepackage{amssymb}

\begin{document}

\title{Not All Agents Matter: From Global Attention Dilution to Risk-Prioritized Game Planning} 

\titlerunning{GameAD}

\author{Kang Ding\inst{1} \and
Hongsong Wang\inst{2} \and
Jie Gui\inst{1} \and Lei He\inst{3}}

\authorrunning{K.~Ding et al.}

\institute{School of Cyberspace Security, Southeast University \and
School of Computer Science and Engineering, Southeast University \and
Vehicle and Mobility School, Tsinghua University}

\maketitle

\begin{abstract}
    End-to-end autonomous driving resides not in the integration of perception and planning, but rather in the dynamic multi-agent game within a unified representation space. Most existing end-to-end models treat all agents equally, hindering the decoupling of real collision threats from complex backgrounds. To address this issue, We introduce the concept of Risk-Prioritized Game Planning, and propose GameAD, a novel framework that models end-to-end autonomous driving as a risk-aware game problem. The GameAD integrates Risk-Aware Topology Anchoring, Strategic Payload Adapter, Minimax Risk-Aware Sparse Attention, and Risk Consistent Equilibrium Stabilization to enable game theoretic decision making with risk prioritized interactions. We also present the Planning Risk Exposure metric, which quantifies the cumulative risk intensity of planned trajectories over a long horizon for safe autonomous driving. Extensive experiments on the nuScenes and Bench2Drive datasets show that our approach significantly outperforms state-of-the-art methods, especially in terms of trajectory safety.
\end{abstract}

\section{Introduction}
\label{sec:intro}
Autonomous driving\cite{chen2024end,Song2024RobustnessAware3O,Wang2023MultiModal3O} systems have traditionally relied on modular pipelines that decompose the task into perception\cite{Li2022BEVFormerLB,MapTR,Wang2023ExploringOT}, prediction\cite{shi2022motion,Zhou2023QueryCentricTP}, and planning\cite{Cheng2024RethinkingIP,dauner2023parting} stages. Although such architectures provide clear functional decomposition, errors may accumulate across stages and limit overall system performance. End-to-end methods\cite{hu2023_uniad,DBLP:conf/iccv/JiangCXLCZZ0HW23,Prakash2021MultiModalFT} aim to learn a unified mapping from sensor observations to driving actions, allowing joint optimization of perception and planning. This paradigm improves representation sharing across tasks and has emerged as a promising direction for scalable autonomous driving systems.

\begin{figure}[tb]
  \centering
  \begin{subfigure}[t]{0.46\linewidth}
  \centering
    \includegraphics[width=\linewidth]{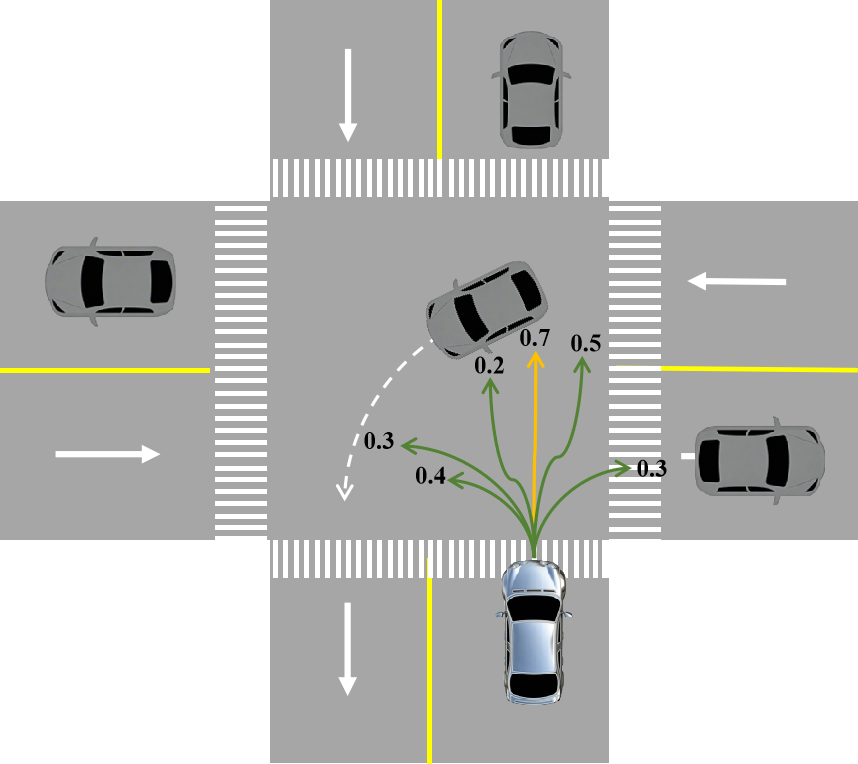}
    \caption{Previous methods}
    \label{fig:a}
  \end{subfigure}
  \hfill
    \begin{subfigure}[t]{0.46\linewidth}
        \centering
        \includegraphics[width=\linewidth]{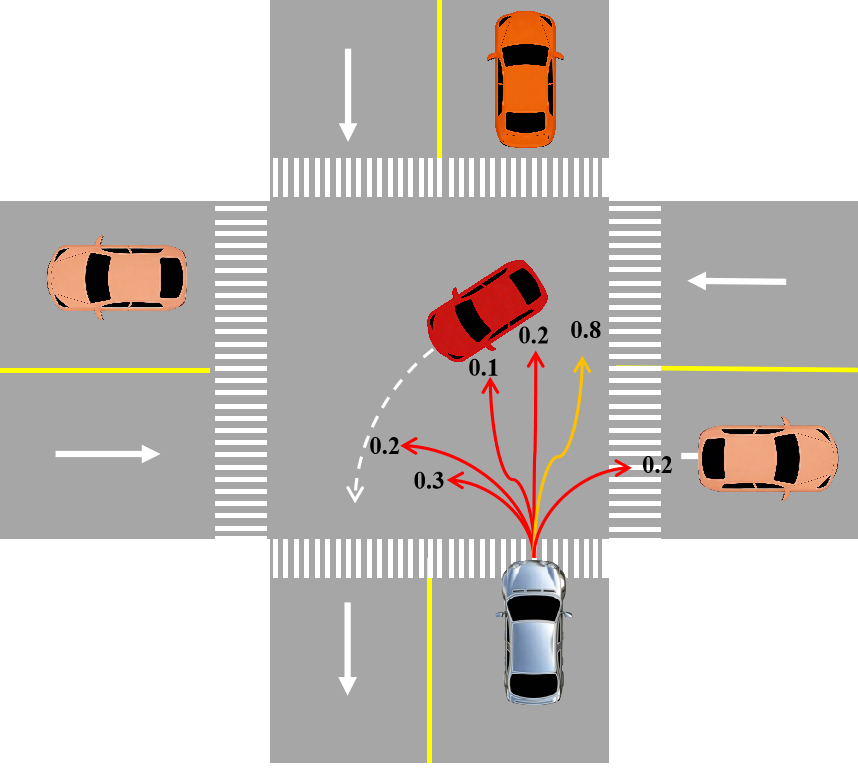}
        \caption{Our GameAD}
        \label{fig:b}
    \end{subfigure}
  \caption{Comparison between our proposed end-to-end autonomous driving paradigm and previous approaches. (a) Previous approaches apply uniform interactions between each plan query and all agents during planning, implicitly assuming that all agents are equally important. (b) Our method computes a collision-risk matrix between each planning mode and agents, which directs attention toward potential conflict agents instead of being diluted by globally irrelevant targets.}
  \label{fig:short}
\end{figure}

Current end-to-end methods\cite{Lin2023Sparse4DVR,Wang2023ExploringOT,Zhang2024SparseADSQ,hu2023_uniad,DBLP:conf/iccv/JiangCXLCZZ0HW23,Huang_2023_ICCV} have made substantial progress with a unified framework. For example, UniAD\cite{hu2023_uniad} formulates driving as a multi-task learning framework that integrates perception and motion prediction to support planning, while VAD\cite{DBLP:conf/iccv/JiangCXLCZZ0HW23} adopts query-based representations to directly retrieve surrounding context for trajectory generation. Despite their effectiveness, most existing approaches treat interactions between the ego vehicle and surrounding agents in a uniform manner, implicitly assuming that all agents contribute equally to decision-making, as illustrated in \cref{fig:short}(a). However, traffic interactions are inherently strategic and can be viewed as a game between the ego vehicle and other participants. To capture such dependencies, GameFormer~\cite{Huang_2023_ICCV} explores game-theoretic formulations and models multi-agent interactions through level-$k$ reasoning, where agents iteratively respond to predicted behaviors of others. Although this mechanism improves behavioral modeling, it does not explicitly evaluate the collision risk posed by different agents and therefore lacks a principled way to prioritize safety-critical participants during planning. Under standard softmax attention, each agent receives approximately $1/K$ attention weight when the number of surrounding agents becomes large, leading to global attention dilution. As a result, planning decisions may overlook agents with geometric conflict, leading to risk-agnostic interaction reasoning and degraded safety under complex multi-agent conditions.

Experienced drivers rapidly prioritize a small number of safety-critical agents while maintaining only coarse awareness of irrelevant vehicles. Inspired by this observation, we reformulate autonomous driving as a risk-aware interaction game and introduce GameAD. As shown in \cref{fig:short}(b), instead of treating interaction modeling as uniform reasoning over all agents, GameAD explicitly models planning as a minimax decision process in which the ego vehicle selects strategies under worst-case opponent behaviors. 
Our contributions are as follows:
\begin{itemize}
    \item \textbf{Risk-Prioritized Game Planning.} We reformulate end-to-end autonomous driving as a risk-conditioned game problem, drawing an analogy to human selective attention in threat-aware driving. We show that standard cross-attention can be viewed as a special case of our formulation when the risk prior is removed, which provides an interpretable perspective on why risk-prioritized sparse game helps improve planning safety in multi-agent scenarios.
    \item \textbf{GameAD Framework.} We propose GameAD, an end-to-end autonomous driving framework based on risk-prioritized game planning. The framework combines Risk-Aware Topology Anchoring, Strategic Payload Adapter, Minimax Risk-Aware Sparse Attention, and Risk-Consistent Equilibrium Stabilization to propagate collision risk information from perception to planning, improving safety and trajectory consistency in multi-agent interactions. 
    \item \textbf{Planning Risk Exposure Metric.} We propose Planning Risk Exposure (PRE) metric, which first introduces spatiotemporal integration of risk exposure into autonomous driving planning evaluation and quantifies the cumulative risk intensity of a planned trajectory over the full time horizon.
    \item \textbf{Performance Evaluation.} Extensive experiments conducted on the nuScenes and Bench2Drive datasets demonstrate that GameAD significantly outperforms state-of-the-art end-to-end autonomous driving planning methods in terms of trajectory safety. 
\end{itemize}

\section{Related Works}
\subsection{Perception for Autonomous Driving}
\label{sec:perception}
Perception aims to understand the surrounding environment by 3D detection, multi-object tracking, and online mapping from multi-sensor inputs. For 3D detection, LSS\cite{philion2020lift} lifts multi-camera image features into 3D frustums and splats them onto a unified bird's-eye-view representation for robust multi-view fusion and planning-oriented scene understanding. Follow-up works\cite{liu2022bevfusion,huang2021bevdet,huang2022bevdet4d} extend BEV perception by improving multi-modal or temporal feature fusion. Recent methods\cite{Wang2023ExploringOT,DBLP:journals/corr/abs-2211-10581,Wang2023MultiModal3O} model long-term temporal dependencies through efficient sparse feature propagation. For multi-object tracking, some methods\cite{yin2021center,Wang2023MultiModal3O} perform object-centric detection with temporal modeling for consistent dynamic object understanding, while others\cite{zeng2021motr,meinhardt2021trackformer} achieve end-to-end tracking via query-based temporal propagation and association. For online mapping, HDMapNet\cite{DBLP:journals/corr/abs-2107-06307} learns vectorized HD maps online from multi-sensor inputs in bird's-eye-view space and VectorMapNet\cite{liu2022vectormapnet} predicts vectorized HD maps as polylines directly from onboard sensor observations in BEV. MapTR\cite{MapTR} formulates vectorized HD map construction as structured set prediction using a Transformer framework.

\subsection{End-to-End Motion Prediction}
\label{sec:motion prediction}
Motion prediction is critical for autonomous systems to reason about traffic evolution and support decision-making. MTR++\cite{shi2023mtr} predicts multimodal futures via intention queries for efficient joint forecasting, while DeMo\cite{DBLP:conf/nips/ZhangSZ24} disentangles trajectory queries into mode and state representations. ViP3D\cite{vip3d} unifies perception and prediction through query-based video modeling for differentiable vision-centric trajectory prediction. Trajectory Mamba\cite{huang2024trajectory} employs selective state-space modeling for linear-complexity multimodal trajectory prediction with joint polyline encoding and cross-state space interaction. PPT\cite{xu2025ppt} pretrains motion forecasting using pseudo-labeled trajectories from off-the-shelf detection and tracking to improve generalization. 

\subsection{End-to-End Planning}
\label{sec:planning}
End-to-end planning directly maps raw sensor data to driving trajectories through integrated architectures. UniAD\cite{hu2023_uniad} unifies perception, prediction, and planning within a single network using shared query interfaces to reduce error accumulation and enable coordinated task optimization toward planning. VAD\cite{DBLP:conf/iccv/JiangCXLCZZ0HW23} models driving scenes with vectorized representations, enabling instance-level planning constraints, faster inference without rasterization. VADv2\cite{jiang2026vadv} formulates end-to-end planning as probabilistic action modeling by discretizing continuous planning space into tokens interacting with scene representations under demonstration supervision. GenAD\cite{zheng2024genad} formulates end-to-end autonomous driving as generative modeling, using instance-centric tokenization, VAE-based structural latent trajectory learning, and temporal modeling to jointly perform motion prediction and planning. SparseDrive\cite{Sun2024SparseDriveEA} introduces sparse scene representation with symmetric perception and parallel multimodal planning, reducing BEV computation while improving efficiency and planning safety in end-to-end autonomous driving. BridgeAD\cite{zhang2025bridging} reformulates motion and planning as multi-step queries, aligning historical aggregation with future prediction to improve perception and planning coherence in end-to-end autonomous driving. MomAD\cite{Song2025DontST} introduces trajectory and perception momentum with Hausdorff-based matching and temporal interaction, improving long-horizon consistency and robustness for end-to-end autonomous driving on nuScenes. GameFormer~\cite{Huang_2023_ICCV} explores game-theoretic formulations and models multi-agent interactions through level-$k$ reasoning, where agents iteratively respond to predicted behaviors of others. While effective for interaction modeling, such formulations implicitly assume equal participation of all agents in decision making. In contrast, real-world driving decisions are typically dominated by a small subset of safety-critical agents. GameAD addresses this challenge by reformulating end-to-end planning as a risk-conditioned sparse game, where interaction is selectively performed based on collision risk rather than uniformly across agents. This shifts the objective from interaction reasoning to risk-aware decision prioritization, distinguishing GameAD from prior interaction-centric game-based planning frameworks.

\section{Methodology: Risk-Prioritized Game Planning}
\label{method}
\begin{figure}[tb]  
    \centering  \includegraphics[width=\linewidth,keepaspectratio]{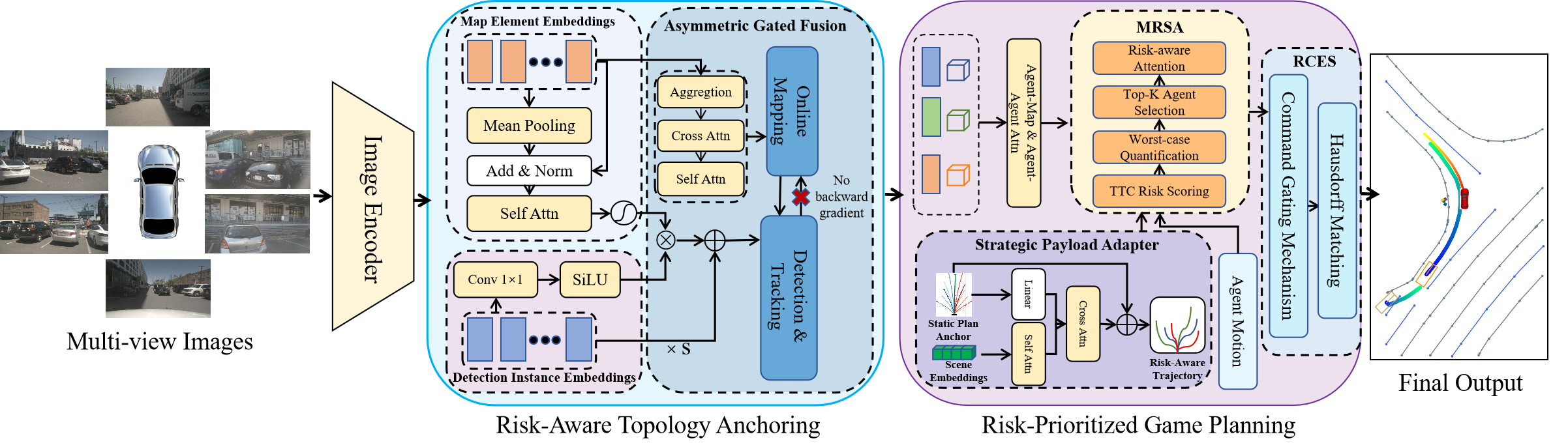}  
    \caption{The overall architecture of GameAD. After multi-scale feature extraction from multi-view images, the framework integrates four core components. Risk-Aware Topology Anchoring transfers road topology information unidirectionally into detection anchors, providing perception features enriched with risk-aware semantics. Strategic Payload Adapter subsequently combines candidate trajectories with perceived scene embeddings through trajectory-scene cross-attention, enabling risk-aware trajectory representation. During decision making, Minimax Risk-aware Sparse Attention(MRSA) computes the worst-case collision risk for each pair of ego planning modes and surrounding agents, guiding attention toward potential conflict participants. Finally, Risk-Consistent Equilibrium Stabilization connects historical and current planning spaces to produce temporally consistent multimodal trajectories.
    }  
    \label{fig:architecture}
\end{figure}

We reformulate end-to-end autonomous driving as a risk-prioritized sparse game problem. Unlike prior frameworks that assign uniform importance to all traffic participants, the core principle of our method is that planning decisions are driven by a small set of safety-critical agents. As shown in \cref{fig:architecture}, our GameAD propagates geometric collision risk through four integrated stages: (1) Risk-Aware Topology Anchoring (RTA), (2) Strategic Payload Adapter (SPA), (3) Minimax Risk-aware Sparse Attention (MRSA), and (4) Risk-Consistent Equilibrium Stabilization (RCES).

\subsection{Risk-Aware Topology Anchoring}
\label{sec:Risk-Aligned Perception}
To enable accurate risk-aware game reasoning, the ego vehicle need to anchor its perception in geometrically sensitive regions. However, existing end-to-end frameworks design detection and mapping as parallel perception branches, which limits performance due to structural information isolation. We propose Risk-Aware Topology Anchoring (RTA) to bridge this gap. RTA moves beyond scene-agnostic anchors by unidirectionally injecting online reconstructed map topology elements into the detection query space.

Specifically, map element embeddings $E_{\text{map}} \in \mathbb{R}^{N_{\text{map}} \times C}$ are first compressed into a scene-level semantic embedding $\bar{E}_{\text{map}} \in \mathbb{R}^{1 \times C}$ via mean pooling, where $C$ denotes the feature dimension. Subsequently, $\bar{E}_{\text{map}}$ generates a global gating embedding $g \in \mathbb{R}^{1 \times C}$ through a residual connection and self-attention mechanism. This process is formulated as:
\begin{align}
    &\bar{E}_{\text{map}} = \mathrm{MeanPooling}(E_{\text{map}}),\\
    &g = \sigma\!\left(\mathrm{Attention}(\bar{E}_{\text{map}} + E_{\text{map}})\right).
\end{align}      

Meanwhile, detection instance embeddings $E_{\text{det}}$ pass through convolutional layers and activation functions to produce instance-level modulation embeddings $\delta$. Finally, following an asymmetric gated fusion strategy, the global gate $g$ interacts with $\delta$ through element-wise multiplication and residual operations to generate geometrically sensitive detection representations $\bar{E}_{\text{det}}$. This design ensures that the detection network prioritizes participants located within potential conflict zones.

\subsection{Strategic Payload Adapter}
\label{sec:strategy_adapter}
The game-theoretic strategy space of the ego vehicle is defined by its candidate trajectories. To align this strategy set with the current risk landscape, we introduce the Strategic Payload Adapter (SPA). SPA transforms a static template set into a scene-conditioned differentiable function. By encoding ego motion, agent distribution, and map constraints into a unified context, SPA refines planning anchors to better encompass feasible evasive maneuvers.

Specifically, we construct three types of tokens, including the ego motion token $\mathbf{M}_{\text{ego}} \in \mathbb{R}^{1 \times D}$, top-$K$ detection instance features $\mathbf{F}_{\text{det}} \in \mathbb{R}^{K \times D}$, and top-$K'$ map instance features $\mathbf{F}_{\text{map}} \in \mathbb{R}^{K' \times D}$. These tokens encode ego speed and yaw, the spatial distribution of surrounding agents, and geometric road constraints, respectively. They are concatenated into a unified scene sequence $\mathcal{C} = [\mathbf{S}_{\text{ego}}; \mathbf{F}_{\text{det}}; \mathbf{F}_{\text{map}}] \in \mathbb{R}^{B \times N \times D}$ and processed by self-attention to enable cross-type interaction in a shared representation space. Meanwhile, each static template trajectory $T_{p}$ is linearly projected into the $D$-dimensional space to form a trajectory query $Q_{p}$, which then attends to $\mathcal{C}$ via trajectory-to-scene cross-attention, ensuring that each candidate is geometrically valid and scene-aware. To preserve stability in early training, a residual addition strategy is applied to the risk-aware trajectory representations. The above process is described as:
\begin{equation}
\bar{T}_{p} = \mathrm{Decoder}\big(\mathrm{CrossAttention}(Q=\mathrm{Linear}(T_{p}), K=\mathcal{C})\big) + T_{p}.
\end{equation}

\subsection{Minimax Risk-Aware Sparse Attention}
\label{sec:mrsa}
\begin{figure}[tb]      
    \centering  \includegraphics[width=\linewidth,keepaspectratio]{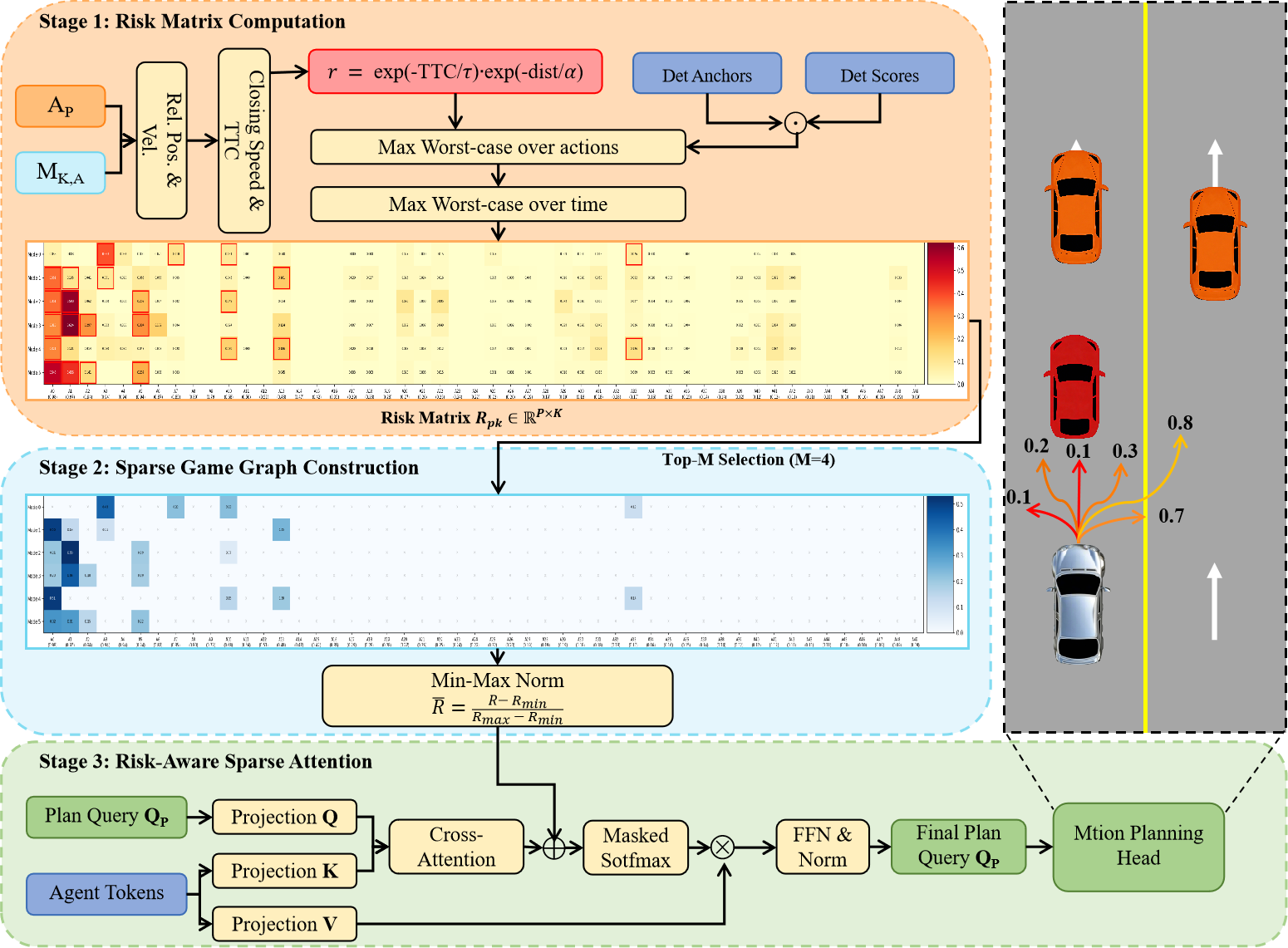}      
    \caption{The architecture of the Minimax Risk-aware Sparse Attention (MRSA) module. In the first stage, Risk Matrix Computation, the framework evaluates the worst-case geometric collision risk by analyzing the motion relationship between the ego planning trajectories and surrounding agents, thereby generating an ego-agent risk matrix. The second stage, Sparse Game Graph Construction, filters the risk matrix to identify a small subset of high-risk adversaries for each planning mode, which establishes a sparse game interaction structure. In the third stage, Risk-Aware Sparse Attention, the normalized risk values are injected into the attention mechanism as an adversarial prior. This process ensures that the planning queries concentrate on safety-critical agents and facilitates risk-aware interactive reasoning.  
    }      
    \label{fig:mrsa}
\end{figure}
Standard cross-attention suffers from global attention dilution, where safety-critical threats are obscured by irrelevant background agents.To address this limitation, we formally model the interaction between the ego vehicle and surrounding agents as a minimax decision process. The minimax risk formulation evaluates the worst-case collision risk for each ego planning mode $p$ and agent $k$ by maximizing over the predicted motion modes $a$ of agent $k$ and prediction time steps $t$. This maximization reflects an adversarial assumption in which the planner accounts for the most dangerous feasible behavior of other agents.

As illustrated in \cref{fig:mrsa}, in the first stage we compute relative position $\boldsymbol{p}_{\mathrm{rel}}$, relative velocity $\boldsymbol{v}_{\mathrm{rel}}$, and time-to-collision (TTC) between planning anchors A$_{P}$ and motion anchors M$_{K,A}$, where $P$ denotes the number of ego planning modes, and $K$ and $A$ denote the number of agents and their motion modes, respectively. The TTC is computed as: 
\begin{align}
\boldsymbol{p}_{\mathrm{rel}} &= \boldsymbol{p}_{\mathrm{agent}} - \boldsymbol{p}_{\mathrm{ego}},\\
\boldsymbol{v}_{\mathrm{rel}} &= \boldsymbol{v}_{\mathrm{agent}} - \boldsymbol{v}_{\mathrm{ego}}, \\
\mathrm{TTC} &= \min\big(\frac{distance}{\max(0,-\frac{\boldsymbol{p}_{\mathrm{rel}} \cdot \boldsymbol{v}_{\mathrm{rel}}}{\|\boldsymbol{p}_{\mathrm{rel}}\|}) + \epsilon},\sigma\big),
\end{align}
where $\boldsymbol{p}_{\mathrm{agent}}$ and $\boldsymbol{p}_{\mathrm{ego}}$ denote the positions of the agent and ego vehicle, respectively; $\boldsymbol{v}_{\mathrm{agent}}$ and $\boldsymbol{v}_{\mathrm{ego}}$ represent the velocities of the agent and ego vehicle, respectively; $distance$ refers to the Euclidean distance between the ego vehicle and the agent; the hyperparameters $\epsilon$ and $\sigma$ are set to $10^{-3}$ and 8.0, respectively.
A joint exponential function combines TTC and spatial distance to produce a risk value $r \in \mathbb{R}^{P \times K \times A \times T}$, followed by maximization over motion modes and temporal steps, corresponding to worst-case opponent behavior and the most critical collision moment. Detection anchors are further weighted by detection confidence scores to obtain the risk matrix $R_{pk}$. The risk matrix is defined as:
\begin{align}
&R_{pk} = \max_{a \in \{1,\dots,A\}} \max_{t \in \{1,\dots,T\}} r,\\
&R_{pk} \leftarrow R_{pk} \cdot \mathrm{clamp}(s_k, 0, 1),
\end{align}
where $s_k$ denotes the detection confidence for agent $k$.

In the second stage, we perform top-M agent selection such that, for each ego mode $p$, only the M agents with the highest risk values are retained. A sparse interaction mask is constructed accordingly, where only top-M entries remain active. The resulting risk matrix is then normalized using min-max normalization to map risk values into the range $[0,1]$.

In the third stage, planning queries $Q_{P}$ attend to the selected high-risk agent tokens through cross-attention, while the normalized risk matrix is injected as an adversarial risk prior into the attention logits, producing the refined planning queries $\bar{Q}_{P}$. Mathematically, $\bar{Q}_{P}$ is computed as:
\begin{align}
    \mathrm{logits_{risk}} = \frac{\mathbf{Q} \cdot \mathbf{K}^T}{\sqrt{d_k}} + \beta \cdot \bar{R}_{pk},\\
\alpha = \sum_k \mathrm{softmax}(\mathrm{logits_{risk}}) \cdot \mathbf{V},\\
   \bar{Q}_{P} = \mathrm{FFN}(\mathrm{LayerNorm}(\alpha)).
\end{align}

This formulation interprets interaction as sparse game reasoning, in which the ego vehicle selectively responds to the most threatening agents, thereby effectively mitigating attention dilution.

\subsection{Risk-Consistent Equilibrium Stabilization}
\label{sec:temporal_fusion}
Frame-wise planning often suffers from strategy oscillation when multiple interaction responses have similar risk levels. To alleviate this issue, we introduce Risk-Consistent Equilibrium Stabilization, which encourages temporal consistency between the current best response and historical strategies. Specifically, we align the current trajectory proposal with historical strategies using the Hausdorff distance, which provides robust global matching between trajectories and is less sensitive to local deviations than point-wise metrics. In addition, the ego command is used as an intention-aware gating signal. When the high-level command changes, historical trajectories are ignored to avoid enforcing outdated strategic priors on new behaviors, ensuring that risk-prioritized decisions evolve into a stable Nash equilibrium over time.

\section{Experiments}
\subsection{Dataset and Evaluation Metrics}
\subsubsection{Dataset.} Experiments are conducted on the challenging nuScenes\cite{caesar2020nuscenes} dataset, which contains 1000 complex driving scenes, of which 700 and 150 sequences are used for training and validation, respectively. Each scene spans approximately 20 seconds. The dataset provides semantic maps and 3D object detection annotations for keyframes sampled at 2 Hz, and each keyframe contains six synchronized camera images. Following established protocols, open-loop evaluation is performed on nuScenes, while closed-loop evaluation is conducted on Bench2Drive\cite{jia2024bench2drive}. Bench2Drive serves as a closed-loop evaluation protocol for end-to-end autonomous driving within the CARLA Leaderboard 2.0 framework. To ensure fair comparison with baseline methods, the official training set is adopted, including the base split consisting of 1000 video clips. Evaluation is performed using the official benchmark comprising 220 routes.
\subsubsection{Evaluation Metrics.} For planning evaluation, we employ the widely used the L2 Displacement Error and the Collision Rate\cite{Hu2022STP3EV} metric to measure planning performance. However, existing mainstream planning metrics primarily focus on binary safety at the outcome level, which fails to account for the duration of the ego vehicle stays within hazardous regions throughout the entire trajectory. To address this limitation, we introduce a novel metric, Planning Risk Exposure (PRE), to quantify the cumulative risk intensity of the planned trajectory across the entire temporal horizon. For each planning time step $t \in \{1, \dots, T\}$, an instantaneous risk value $\Phi(t)$ is computed using ground-truth obstacle boxes, which is defined as:
\begin{equation}
\Phi(t) = \max_{k \in \text{agents}_{t}} \big( \exp(-\frac{\text{TTC}_{kt}}{\tau}) \cdot \exp(-\frac{d_{kt}}{\sigma}) \big),
\end{equation}
where $d_{kt}$ denotes the Euclidean distance between the ego vehicle and obstacle $k$ at time step $t$, while $\tau$ and $\sigma$ serve as decay parameters. The PRE is then obtained by averaging $\Phi(t)$ over the temporal horizon:
\begin{equation}
\text{PRE} = \frac{1}{T} \sum_{t=1}^{T} \Phi(t) \in [0, 1).
\end{equation}
A lower PRE value indicates higher safety, whereas a PRE close to $1$ signifies that the vehicle remains under extreme risk throughout the entire duration.
\subsection{Implementation Details}
Consistent with SparseDrive\cite{Sun2024SparseDriveEA}, ResNet50\cite{He2015DeepRL} is adopted as the backbone for image feature encoding with an input resolution of $256 \times 704$. Training employs the AdamW\cite{Loshchilov2017DecoupledWD} optimizer with Cosine Annealing\cite{loshchilov2017sgdr}, a weight decay of $1 \times 10^{-3}$, and an initial learning rate of $1 \times 10^{-4}$. The training process consists of two stages: the first stage focuses on perception tasks and the second stage performs end-to-end training. All experiments are executed on 4 NVIDIA Tesla A100 GPUs. Additional implementation details and experiments are provided in the supplementary material.

\begin{table}[!t]
  \caption{Open-loop planning results on the nuScenes validation dataset. $^{\dagger}$ denotes evaluation with the official checkpoint. PRE denotes Planning Risk Exposure metric. $^*$ refers the re-implementation. As Ref.\cite{DBLP:conf/cvpr/LiYLLKL024} states, we do not use the ego status information for a fair comparison.
  }
  \label{tab:open-loop planning}
  \centering
  \resizebox{\linewidth}{!}{%
  \begin{tabular}{@{}l c c c c c c c c c c c c c c c}
    \toprule
    \multirow{2}{*}{Method} & \multirow{2}{*}{Reference} & \multirow{2}{*}{Backbone}
    & \multicolumn{4}{c}{$\mathrm{L2}$ (m)$\downarrow$}
    & \multicolumn{4}{c}{Col. Rate (\%)$\downarrow$}
    & \multicolumn{4}{c}{PRE (\%)$\downarrow$}
    & \multirow{2}{*}{FPS$\uparrow$} \\
    \cmidrule(lr){4-7}\cmidrule(lr){8-11}\cmidrule(lr){12-15}
    & & &
    1s & 2s & 3s & \cellcolor{gray!15}Avg.
    & 1s & 2s & 3s & \cellcolor{gray!15}Avg.
    & 1s & 2s & 3s & \cellcolor{gray!15}Avg.
    & \\
    \midrule
    UniAD$^{\dagger}$\cite{hu2023_uniad}      & CVPR2023 & ResNet101
    & 0.45 & 0.70 & 1.04 & \cellcolor{gray!15}0.73
    & 0.62 & 0.58 & 0.63 & \cellcolor{gray!15}0.61
    & - & - & - & \cellcolor{gray!15}-
    & 1.8 (A100) \\
    
    VAD$^{\dagger}$\cite{jiang2026vadv}         & ICCV2023 & ResNet50
    & 0.41 & 0.70 & 1.05 & \cellcolor{gray!15}0.72
    & 0.03 & 0.19 & 0.43 & \cellcolor{gray!15}0.21
    & - & - & - & \cellcolor{gray!15}-
    & 4.5 (RTX3090) \\

    GenAD\cite{zheng2024genad}     & ECCV2024 & ResNet50
    & 0.36 & 0.83 & 1.55 & \cellcolor{gray!15}0.91
    & 0.06 & 0.23 & 1.00 & \cellcolor{gray!15}0.43
    & - & - & - & \cellcolor{gray!15}-
    & 6.7 (RTX3090) \\
    
    SparseDrive$^{\dagger}$\cite{Sun2024SparseDriveEA}& ICRA2025 & ResNet50
    & 0.30 & 0.58 & 0.96 & \cellcolor{gray!15}0.61
    & 0.01 & 0.05 & 0.23 & \cellcolor{gray!15}0.10
    & 4.35 & 4.50 & 4.61 & \cellcolor{gray!15}4.49
    & 7.4 (A100) \\
    
    MomAD\cite{Song2025DontST}        & CVPR2025 & ResNet50
    & 0.31 & 0.57 & 0.91 & \cellcolor{gray!15}0.60
    & 0.01 & 0.05 & 0.22 & \cellcolor{gray!15}0.09
    & 4.33 & 4.41 & 4.52 & \cellcolor{gray!15}4.42
    & 6.3 (A100) \\

    BridgeAD$^*$\cite{zhang2025bridging}       & CVPR2025 & ResNet50    
    & 0.30 & 0.58 & 0.92 & \cellcolor{gray!15}0.60    
    & 0.01 & 0.05 & 0.22 & \cellcolor{gray!15}0.09    
    & \bf3.88 & 3.94 & 4.05 & \cellcolor{gray!15}3.96    
    & 4.4 (A100) \\
    \rowcolor{gray!15}
    GameAD        & - & ResNet50    
    & \textbf{0.30} & \textbf{0.57} & \textbf{0.90} & \cellcolor{gray!15}\textbf{0.59}    
    & \textbf{0.01} & \textbf{0.04} & \textbf{0.18} & \cellcolor{gray!15}\textbf{0.08}    
    & 3.89 & \textbf{3.93} & \textbf{4.01} & \cellcolor{gray!15}\textbf{3.94}    
    &  5.2 (A100) \\
  \bottomrule
  \end{tabular}%
  }
\end{table}
\begin{table}[!t]
    \renewcommand{\arraystretch}{0.85}
    \caption{Open-loop and Closed-loop results on Bench2Drive(V0.0.3) under base training set. ‘mmt’ refers multi-modal trajectory variant of VAD.}
    \label{tab:bench2drive_results}
    \centering
    {\fontsize{8}{9}\selectfont
    \resizebox{0.6\linewidth}{!}{%
    \begin{tabular}{@{}l c c c c c@{}}
    \toprule
    \multirow{2}{*}{Method}
    & Open-loop Metric
    & \multicolumn{4}{c}{Closed-loop Metric} \\
    \cmidrule(lr){2-2}\cmidrule(lr){3-6}
    & Avg.\ L2$\downarrow$
    & DS$\uparrow$
    & SR(\%)$\uparrow$
    & Effi$\uparrow$
    & Comf$\uparrow$ \\
    \midrule
    
    VAD\cite{jiang2026vadv}
    & 0.91 & 42.35 & 15.00 & 157.94 & 46.01 \\

    VAD$_{\text{mmt}}$\cite{jiang2026vadv}
    & 0.89 & 42.87 & 15.91 & 158.12 & 47.22 \\

    GenAD\cite{zheng2024genad}    
    & - & 44.81 & 15.90 & - & - \\
    
    SparseDrive\cite{Sun2024SparseDriveEA}
    & 0.87 & 44.54 & 16.71 & 170.21 & 48.63 \\
    
    MomAD\cite{Song2025DontST}
    & 0.82 & 47.91 & 18.11 & 174.91 & 51.20 \\
    \rowcolor{gray!15}
    GameAD
    & \bf0.80 & \bf48.35 & \bf19.43 & \bf178.96 & \bf53.11 \\
    
    \bottomrule
    \end{tabular}%
    }
    }
\end{table}

\subsection{Main Results}
\subsubsection{Planning Results on nuScenes.}
As illustrated in \cref{tab:open-loop planning}, our GameAD is evaluated against recent state-of-the-art end-to-end autonomous driving frameworks. GameAD establishes a new performance benchmark on the nuScenes dataset. Specifically, our proposed method attains an average L2 error of $0.59$m, matching the performance of MomAD and remains highly competitive with BridgeAD. Regarding safety metrics, GameAD achieves the lowest average collision rate of $0.08\%$, representing a $11\%$ reduction relative to BridgeAD. Furthermore, GameAD outperforms MomAD by $11\%$ in terms of the PRE metric. These results indicate that GameAD further strengthens planning safety.
\subsubsection{Planning Results on Bench2Drive.}
As shown in \cref{tab:bench2drive_results}, we evaluate closed-loop performance on the Bench2Drive dataset. GameAD improves the success rate by $16.3\%$ and $7.3\%$ relative to SparseDrive and MomAD, respectively, highlighting the importance of risk awareness for safe driving. Simultaneously, GameAD achieves an Efficiency metric of 178.96, surpassing both MomAD at 174.91 and SparseDrive at 170.21. This improvement in efficiency demonstrates that risk-prioritized game-theoretic reasoning does not lead to overly conservative driving behavior.
\subsubsection{Perception and Motion Prediction Results.}
The parallel design of 3D detection and online mapping in SparseDrive\cite{Sun2024SparseDriveEA} leads to structural information isolation. To address this issue, we injects the reconstructed road topology into the anchor space of the detection branch to strengthen perception. As shown in \cref{tab:multi_task_results}, our GameAD achieves a 3D detection mAP of $0.425$ and an NDS of $0.536$ on nuScenes, improving over SparseDrive by $1.7\%$ in mAP and $2.1\%$ in NDS. In terms of multi-object tracking, GameAD reaches an AMOTA of $39.9\%$ with $566$ IDS, outperforming MomAD by $2\%$ in AMOTA and reducing IDS by $33.6\%$. In terms of online mapping, GameAD achieves $57.2\%$ mAP, which is $2.3\%$ higher than MomAD. In terms of motion prediction, GameAD outperforms both MomAD and BridgeAD, achieving superior prediction performance.

\begin{table}[!t]
    \caption{Perception and motion results on the nuScenes validation dataset. $^{\dagger}$ denotes evaluation with the official checkpoint. AP$_{d}$ denotes AP$_{divider}$. AP$_{b}$ denotes AP$_{boundary}$. mADE denotes minADE. mFDE denotes minFDE.}
    \label{tab:multi_task_results}
    \centering
    \resizebox{\linewidth}{!}{%
    \begin{tabular}{@{}l c c c c c c c c c c c c c c@{}}
    \toprule
    \multirow{2}{*}{Method}
    & \multicolumn{2}{c}{3D Object Detection}
    & \multicolumn{4}{c}{Multi-Object Tracking}
    & \multicolumn{4}{c}{Online Mapping}
    & \multicolumn{4}{c}{Motion Prediction} \\
    \cmidrule(lr){2-3}\cmidrule(lr){4-7}\cmidrule(lr){8-11}\cmidrule(lr){12-15}
    & mAP$\uparrow$ & NDS$\uparrow$
    & AMOTA$\uparrow$ & AMOTP$\downarrow$ & Recall$\uparrow$ & IDS$\downarrow$
    & mAP$\uparrow$ & AP$_{\text{ped}}$$\uparrow$ & AP$_d$$\uparrow$ & AP$_b$$\uparrow$
    & mADE$\downarrow$ & mFDE$\downarrow$ & MR$\downarrow$ & EPA$\uparrow$ \\
    \midrule
    UniAD\cite{hu2023_uniad}
    & 0.380 & 0.498
    & 0.359 & 1.320 & 0.467 & 906
    & - & - & - & -
    & 0.71 & 1.02 & 0.151 & 0.456 \\
    VAD$^{\dagger}$\cite{jiang2026vadv}
    & 0.312 & 0.435
    & - & - & - & -
    & 47.6 & 40.6 & 51.5 & 50.6
    & - & - & - & - \\
    SparseDrive\cite{Sun2024SparseDriveEA}
    & 0.418 & 0.525
    & 0.386 & 1.254 & 0.499 & 886
    & 55.1 & 49.9 & 57.0 & 58.4
    & 0.62 & 0.99 & 0.136 & 0.482 \\
    MomAD\cite{Song2025DontST}
    & 0.423 & 0.531
    & 0.391 & 1.243 & 0.509 & 853
    & 55.9 & 50.7 & 58.1 & \textbf{58.9}
    & 0.61 & 0.98 & 0.137 & 0.499 \\
    BridgeAD\cite{zhang2025bridging}    
    & 0.423 & 0.534   
    & 0.398 & 1.232 & 0.501 & 639    
    & - & - & - & -    
    & 0.62 & 0.98 & 0.134 & 0.504 \\
    \rowcolor{gray!15}
    GameAD    
    & \textbf{0.425} & \textbf{0.536}    
    & \textbf{0.399} & \textbf{1.224} & \textbf{0.504} & \textbf{566}    
    & \textbf{57.1} & \textbf{53.6} & \textbf{59.1} & 58.7    
    & \textbf{0.60} & \textbf{0.97} & \textbf{0.132} & \textbf{0.507} \\
    \bottomrule
    \end{tabular}%
    }
\end{table}

\begin{table}[!t]
    \caption{Ablation studies of the risk game planning on the nuScenes validation split. Distance denotes the metric used to measure trajectory consistency. $risk_\beta$ represents the factor that controls risk intensity, and $t$ denotes the number of fused historical frames. TPC denotes Trajectory Prediction Consistency metric in \cite{Song2025DontST}.}
    \label{tab:charactor_ablation}
    \centering
    \resizebox{\linewidth}{!}{%
    \begin{tabular}{@{}c c c c c c c c c c c c c c c c@{}}
    \toprule
    \multirow{2}{*}{ID}
    & \multirow{2}{*}{Distance}
    & \multirow{2}{*}{$risk_\beta$}
    & \multirow{2}{*}{$t$}
    & \multicolumn{4}{c}{$\mathrm{L2}$ (m)$\downarrow$}
    & \multicolumn{4}{c}{Col. Rate (\%)$\downarrow$}
    & \multicolumn{4}{c}{TPC (m)$\downarrow$} \\
    \cmidrule(lr){5-8}\cmidrule(lr){9-12}\cmidrule(lr){13-16}
    & & & &
    1s & 2s & 3s & Avg.
    & 1s & 2s & 3s & Avg.
    & 1s & 2s & 3s & Avg. \\
    \midrule
    
    0 & Euclidean & 1.0 & 1
    & 0.350 & 0.672 & 1.091 & 0.704
    & 0.053 & 0.102 & 0.256 & 0.137
    & 0.302 & 0.524 & 0.769 & 0.532 \\
    
    1 & Hausdorff & 1.0 & 1
    & 0.341 & 0.661 & 1.082 & 0.695
    & 0.049 & 0.098 & 0.251 & 0.133
    & 0.296 & 0.519 & 0.763 & 0.526 \\
    
    2 & Hausdorff & 2.0 & 1
    & 0.326 & 0.622 & 1.000 & 0.649
    & 0.010 & 0.059 & 0.195 & 0.088
    & 0.287 & \textbf{0.501} & \textbf{0.736} & \textbf{0.508} \\
    
    3 & Hausdorff & 4.0 & 1
    & 0.331 & 0.654 & 1.082 & 0.689
    & 0.059 & 0.088 & 0.247 & 0.131
    & 0.287 & 0.507 & 0.753 & 0.516 \\
    
    5 & Hausdorff & 2.0 & 2
    & 0.321 & 0.614 & 0.999 & 0.645
    & 0.020 & 0.054 & 0.234 & 0.103
    & 0.298 & 0.519 & 0.759 & 0.525 \\
    \rowcolor{gray!15}
    6 & Hausdorff & 2.0 & 3
    & \textbf{0.304} & \textbf{0.572} & \textbf{0.904} & \textbf{0.593}
    & \textbf{0.010} & \textbf{0.039} & \textbf{0.182} & \textbf{0.077}
    & \textbf{0.278} & 0.504 & 0.744 & 0.509 \\

    7 & Hausdorff & 2.0 & 4    
    & 0.314 & 0.603 & 0.987 & 0.635    
    & 0.020 & 0.047 & 0.212 & 0.093    
    & 0.290 & 0.511 & 0.751 & 0.517 \\
    
    \bottomrule
    \end{tabular}%
    }
\end{table}

\begin{table}[!t]
    \caption{Ablation studies of the core modules in GameAD on the nuScenes validation split. RTA represents Risk-Aware Topology Anchoring. SPA represents Strategic Payload Adapter. MRSA represents Minimax Risk-Aware Sparse Attention. RCES represents Risk-Consistent Equilibrium Stabilization.}
    \label{tab:module_ablation}
    \centering
    \resizebox{\linewidth}{!}{%
    \begin{tabular}{@{}c c c c c c c c c c c c c c c c c c@{}}
    \toprule
    \multirow{2}{*}{ID}
    & \multirow{2}{*}{RTA}
    & \multirow{2}{*}{MRSA}
    & \multirow{2}{*}{RCES}
    & \multirow{2}{*}{SPA}
    & \multicolumn{4}{c}{PRE (\%)$\downarrow$}
    & \multicolumn{4}{c}{Coll. Rate (\%)$\downarrow$}
    & \multicolumn{4}{c}{TPC (m)$\downarrow$} \\
    
    \cmidrule(lr){6-9}
    \cmidrule(lr){10-13}
    \cmidrule(lr){14-17}
    
    & & & & 
    & 1s & 2s & 3s & Avg.
    & 1s & 2s & 3s & Avg.
    & 1s & 2s & 3s & Avg. \\
    \midrule
    
    0 & \checkmark & \checkmark & \checkmark & \checkmark
    & \textbf{3.89} & \textbf{3.93} & \textbf{4.01} & \textbf{3.94}
    & \textbf{0.01} & \textbf{0.04} & \textbf{0.18} & \textbf{0.08}
    & \textbf{0.28} & \textbf{0.50} & \textbf{0.74} & \textbf{0.51} \\
    
    1 &  & \checkmark & \checkmark & \checkmark
    & 3.92 & 3.97 & 4.05 & 3.98
    & 0.01 & 0.07 & 0.22 & 0.10
    & 0.28 & 0.51 & 0.75 & 0.51 \\
    
    2 & \checkmark &  & \checkmark & \checkmark
    & 3.95 & 4.01 & 4.10 & 4.02
    & 0.02 & 0.10 & 0.33 & 0.15
    & 0.30 & 0.52 & 0.76 & 0.53 \\
    
    3 & \checkmark & \checkmark &  & \checkmark
    & 3.90 & 3.95 & 4.02 & 3.96
    & 0.02 & 0.08 & 0.26 & 0.12
    & 0.31 & 0.54 & 0.77 & 0.54 \\
    
    4 & \checkmark & \checkmark & \checkmark &
    & 3.91 & 3.99 & 4.04 & 3.98
    & 0.01 & 0.08 & 0.29 & 0.13
    & 0.29 & 0.51 & 0.75 & 0.52 \\
    
    \bottomrule
    \end{tabular}%
    }
\end{table}
\begin{table}[!t]
    \caption{Ablation studies of perception. AP$_{d}$ denotes AP$_{divider}$. AP$_{b}$ denotes AP$_{boundary}$. ‘perception’ follows the setting of SparseDrive\cite{Sun2024SparseDriveEA}. ‘instance robust’ is re-implemented based on MomAD\cite{Song2025DontST}. ‘motion2det’ is re-implemented based on BridgeAD\cite{zhang2025bridging}. ‘uni. calibration’ indicates the flow from the map branch to the detection branch, while ‘bid. calibration’ indicates mutual influence between the two branches.}
    \label{tab:perception_ablation}
    \centering
    \resizebox{\linewidth}{!}{%
    \begin{tabular}{@{}c l c c c c c c c c c c@{}}
    \toprule
    \multirow{2}{*}{ID}
    & \multirow{2}{*}{Method}
    & \multicolumn{2}{c}{3D Object Detection}
    & \multicolumn{4}{c}{Multi-Object Tracking}
    & \multicolumn{4}{c}{Online Mapping} \\
    \cmidrule(lr){3-4}\cmidrule(lr){5-8}\cmidrule(lr){9-12}
    & & mAP$\uparrow$ & NDS$\uparrow$
    & AMOTA$\uparrow$ & AMOTP$\downarrow$ & Recall$\uparrow$ & IDS$\downarrow$
    & mAP$\uparrow$ & AP$_{\text{ped}}$$\uparrow$ & AP$_d$$\uparrow$ & AP$_b$$\uparrow$\\
    \midrule
    0 & perception
    & 0.418 & 0.525
    & 0.386 & 1.254 & 0.499 & 886
    & 55.1 & 49.9 & 57.0 & 58.4 \\
    1 & perception $+$ instance robust
    & 0.420 & 0.527
    & 0.390 & 1.248 & \bf0.505 & 864
    & 55.2 & 50.1 & 57.3 & 58.2 \\
    2 & perception $+$ motion2det    
    & 0.421 & \bf0.537    
    & 0.396 & 1.234 & 0.502 & 574    
    & 54.2 & 52.1 & 54.3 & 56.2 \\
    \rowcolor{gray!15}
    3 & perception $+$ uni. calibration 
    & \bf0.425 & 0.536    
    & \bf0.399 & \bf1.224 & 0.504 & \bf566    
    & \bf57.1 & \bf53.6 & \bf59.1 & \bf58.7 \\
    4 & perception $+$ bid. calibration
    & 0.411 & 0.516        
    & 0.371 & 1.273 & 0.453 & 1022        
    & 51.4 & 49.1 & 52.9 & 52.3 \\
    \bottomrule
    \end{tabular}%
    }
\end{table}

\begin{figure}[tb]          
    \centering          
    \includegraphics[width=\linewidth,keepaspectratio]{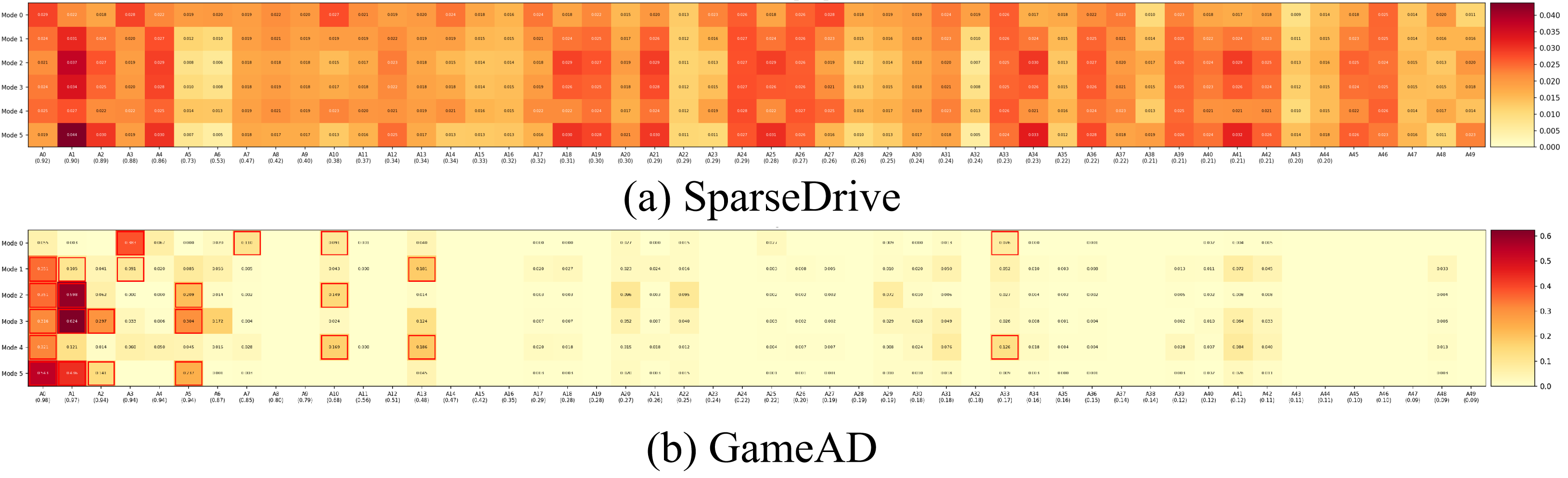}          
    \caption{Visualization of attention heatmaps for the comparison between GameAD and SparseDrive in multi-agent interaction. GameAD exhibits sparse and mode-differentiated interaction structures compared to the dispersed distribution in SparseDrive.    
    }          
    \label{fig:vis_attn}
\end{figure}
\begin{figure}[tb]          
    \centering          
    \includegraphics[width=0.9\linewidth,keepaspectratio]{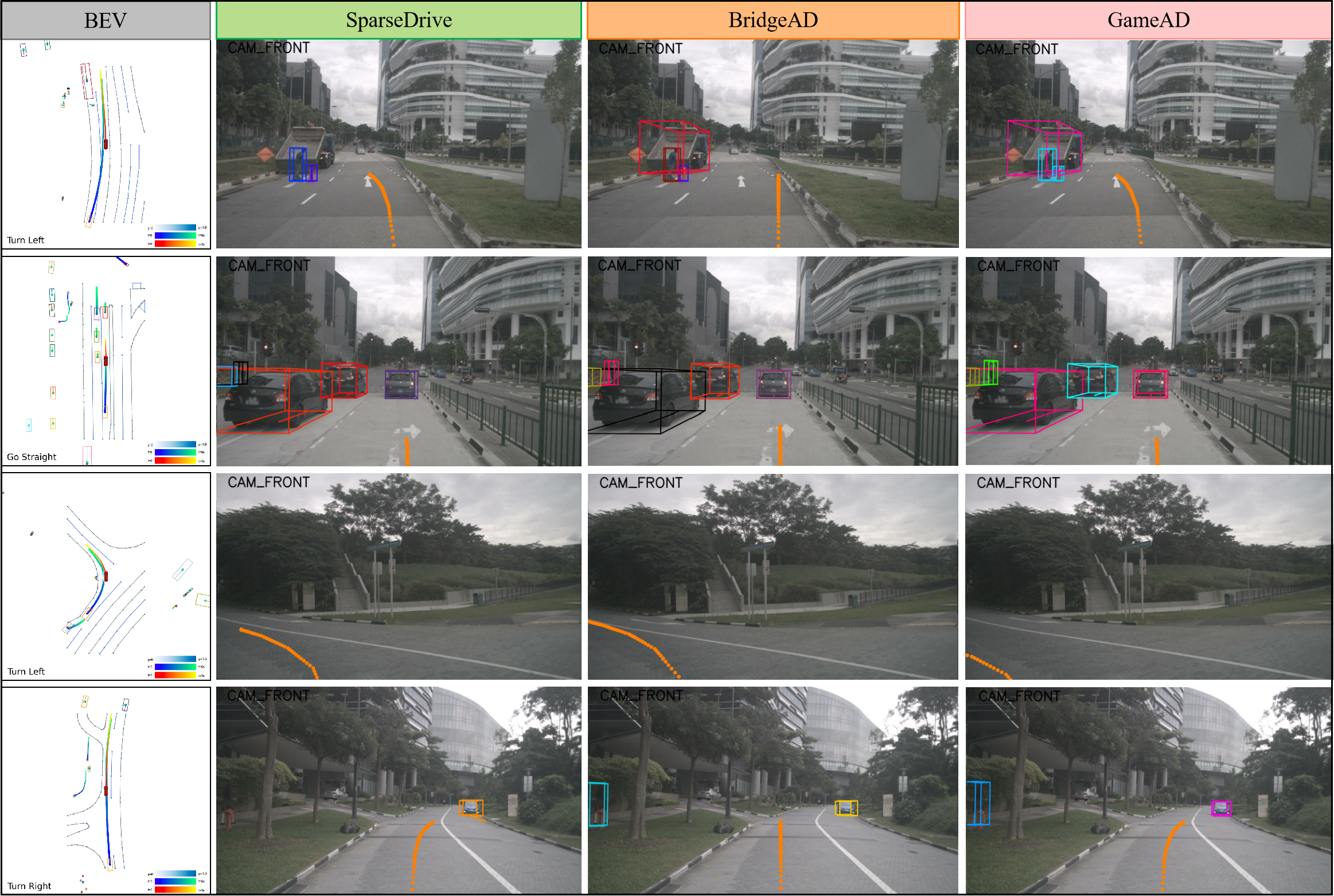}
    \vspace{-2mm}
    \caption{Qualitative comparison of GameAD with SparseDrive and BridgeAD. Our GameAD achieves better performance in both perception and planning.    
    }          
    \label{fig:vis_2}
\end{figure}

\subsection{Ablations and Analysis}

\subsubsection{Impact of Hyperparameter Design on Planning.} As shown in \cref{tab:charactor_ablation}, we analyze the impact of trajectory distance metrics, risk intensity $risk_\beta$, and the number of fused historical frames on planning performance. Under identical settings, replacing Euclidean distance with Hausdorff distance consistently improves all evaluation metrics. We further fix the number of historical frames to $1$ while varying the $risk_\beta$, and then fix the $risk_\beta$ to $2.0$ while varying the number of historical frames. The combination of Hausdorff distance, a $risk_\beta$ of $2.0$, and $3$ historical frames achieves the best performance.
\subsubsection{Impact of Modules Design on Planning.} As shown in \cref{tab:module_ablation}, we conduct ablation studies on the core modules to quantify their contributions to comprehensive performance. From ID-1 to ID-4, we remove RTA, MRSA, RCES, and SPA, respectively. The results show that removing any single module consistently degrades planning performance, indicating that the performance gains of our GameAD stem from the synergy among multiple components rather than from a single dominant factor.
\subsubsection{Impact of Design on Perception.} As shown in \cref{tab:perception_ablation}, we analyze the influence of different components on perception performance. From ID-1 to ID-4, we progressively add instance robust, motion2det, and topology calibration to enhance the baseline SparseDrive\cite{Sun2024SparseDriveEA}. Unidirectional topology calibration achieves the best comprehensive performance in perception. In contrast, bidirectional topology calibration leads to a significant performance degradation. We attribute this degradation to optimization interference introduced by coupling the detection and mapping branches in both directions, whereas a unidirectional design supports more stable multi-task joint learning.

\subsubsection{Efficiency Analysis.}
As shown in \cref{tab:open-loop planning}, we compare the Frames Per Second (FPS) of our GameAD and other end-to-end methods. The FPS for all models is evaluated on either a single NVIDIA RTX 3090 GPU or an NVIDIA Tesla A100 GPU with a batch size of 1. The inference latency of our model is 192.3 ms, faster than BridgeAD's 227.3 ms but slower than MomAD's 158.7 ms. 

\subsubsection{Qualitative Results.}
As illustrated in \cref{fig:vis_attn}, the attention heatmaps reveal a distinct difference between GameAD and SparseDrive in terms of importance allocation during agent interactions. In SparseDrive, attention weights exhibit a dispersed distribution across a large number of agents. Furthermore, the attention patterns in SparseDrive show minimal variation across different planning modes, suggesting a limited capability to distinguish between diverse ego behaviors. In contrast, GameAD produces a sparser and more mode-differentiated interaction structure. Each planning mode concentrates high attention on a small subset of safety-critical agents. This indicates that the Minimax Risk-aware Sparse Attention (MRSA) enables each trajectory mode to selectively identify high-risk agents for refined interactive reasoning, while simultaneously reducing computational redundancy through this selective mechanism.
In \cref{fig:vis_2}, we present a comprehensive visual comparison between GameAD, SparseDrive, and BridgeAD, focusing on both perception and planning performance. While SparseDrive demonstrates reasonable planning capabilities, it exhibits excessive trajectory magnitudes during left-turn maneuvers that introduce collision risks, alongside noticeable degradation in perception accuracy. Conversely, although BridgeAD performs well in perception tasks, its planning module appears to lack sensitivity to road topology. In contrast, the qualitative results indicate that our model, by integrating risk-aware topology anchoring with game-theoretic planning, achieves a precise understanding of surrounding agents. Consequently, GameAD generates reasonable planning trajectories that successfully avoid collisions. Additional qualitative results and failure cases are provided in the supplementary materials.

\section{Conclusion}
We revisit end-to-end autonomous driving from a game-theoretic perspective and identify global attention dilution as a key limitation of unified frameworks, where equal attention to all agents ignores geometrically critical interactions. Planning is reformulated as a risk-conditioned multi-agent game, leading to GameAD, which enables explicit risk propagation from perception to planning. GameAD integrates Risk-Aware Topology Anchoring, Strategic Payload Adapter, Minimax Risk-Aware Sparse Attention, and Risk-Consistent Equilibrium Stabilization. Experiments on nuScenes and Bench2Drive demonstrate state-of-the-art performance in open-loop planning, closed-loop driving, and unified perception tasks. In future work, we plan to introduce the concept of risk-game planning into the research on VLA for autonomous driving.


%
%
\bibliographystyle{splncs04}
\bibliography{main}

@String(CVPR  = {IEEE Conf. Comput. Vis. Pattern Recog.})

@String(ICCV  = {Int. Conf. Comput. Vis.})

@String(ECCV  = {Eur. Conf. Comput. Vis.})

@String(NeurIPS = {Adv. Neural Inform. Process. Syst.})

@String(CVPR  = {CVPR})

@String(ICCV  = {ICCV})

@String(ECCV  = {ECCV})

@String(NeurIPS = {NeurIPS})

@inproceedings{philion2020lift,
    title={Lift, Splat, Shoot: Encoding Images From Arbitrary Camera Rigs by Implicitly Unprojecting to 3D},
    author={Jonah Philion and Sanja Fidler},
    booktitle={Proceedings of the European Conference on Computer Vision},
    year={2020},
}

@inproceedings{liu2022bevfusion,
  title={BEVFusion: Multi-Task Multi-Sensor Fusion with Unified Bird's-Eye View Representation},
  author={Liu, Zhijian and Tang, Haotian and Amini, Alexander and Yang, Xingyu and Mao, Huizi and Rus, Daniela and Han, Song},
  booktitle={IEEE International Conference on Robotics and Automation (ICRA)},
  year={2023}
}

@article{huang2022bevdet4d,
  title={BEVDet4D: Exploit Temporal Cues in Multi-camera 3D Object Detection},
  author={Huang, Junjie and Huang, Guan},
  journal={arXiv preprint arXiv:2203.17054},
  year={2022}
}

@article{huang2021bevdet,
  title={BEVDet: High-performance Multi-camera 3D Object Detection in Bird-Eye-View},
  author={Huang, Junjie and Huang, Guan and Zhu, Zheng and Yun, Ye and Du, Dalong},
  journal={arXiv preprint arXiv:2112.11790},
  year={2021}
}

@article{DBLP:journals/corr/abs-2211-10581,
  publtype={informal},
  author={Xuewu Lin and Tianwei Lin and Zixiang Pei and Lichao Huang and Zhizhong Su},
  title={Sparse4D: Multi-view 3D Object Detection with Sparse Spatial-Temporal Fusion},
  year={2022},
  cdate={1640995200000},
  journal={CoRR},
  volume={abs/2211.10581},
  url={https://doi.org/10.48550/arXiv.2211.10581},
}

@article{yin2021center,
  title={Center-based 3D Object Detection and Tracking},
  author={Yin, Tianwei and Zhou, Xingyi and Kr{\"a}henb{\"u}hl, Philipp},
  journal={CVPR},
  year={2021},
}

@inproceedings{zeng2021motr,
  title={MOTR: End-to-End Multiple-Object Tracking with TRansformer},
  author={Zeng, Fangao and Dong, Bin and Zhang, Yuang and Wang, Tiancai and Zhang, Xiangyu and Wei, Yichen},
  booktitle={European Conference on Computer Vision (ECCV)},
  year={2022}
}

@InProceedings{meinhardt2021trackformer,
    title={TrackFormer: Multi-Object Tracking with Transformers},
    author={Tim Meinhardt and Alexander Kirillov and Laura Leal-Taixe and Christoph Feichtenhofer},
    year={2022},
    month = {June},
    booktitle = {The IEEE Conference on Computer Vision and Pattern Recognition (CVPR)},
}

@article{DBLP:journals/corr/abs-2107-06307,
  publtype={informal},
  author={Qi Li and Yue Wang and Yilun Wang and Hang Zhao},
  title={HDMapNet: An Online HD Map Construction and Evaluation Framework},
  year={2021},
  cdate={1609459200000},
  journal={CoRR},
  volume={abs/2107.06307},
  url={https://arxiv.org/abs/2107.06307}
}

@inproceedings{liu2022vectormapnet,
    title={VectorMapNet: End-to-end Vectorized HD Map Learning},
    author={Liu, Yicheng and Yuan, Tianyuan and Wang, Yue and Wang, Yilun and Zhao, Hang},
    booktitle={International conference on machine learning},
    year={2023},
    organization={PMLR}
}

@inproceedings{MapTR,
  title={MapTR: Structured Modeling and Learning for Online Vectorized HD Map Construction},
  author={Liao, Bencheng and Chen, Shaoyu and Wang, Xinggang and Cheng, Tianheng and Zhang, Qian and Liu, Wenyu and Huang, Chang},
  booktitle={International Conference on Learning Representations},
  year={2023}
}

@article{shi2023mtr,
  title={MTR++: Multi-Agent Motion Prediction with Symmetric Scene Modeling and Guided Intention Querying},
  author={Shi, Shaoshuai and Jiang, Li and Dai, Dengxin and Schiele, Bernt},
  journal={arXiv preprint arXiv:2306.17770},
  year={2023}
}

@inproceedings{vip3d,
  title={ViP3D: End-to-end visual trajectory prediction via 3d agent queries},
  author={Gu, Junru and Hu, Chenxu and Zhang, Tianyuan and Chen, Xuanyao and Wang, Yilun and Wang, Yue and Zhao, Hang},
  booktitle={Proceedings of the IEEE/CVF Conference on Computer Vision and Pattern Recognition},
  pages={5496--5506},
  year={2023}
}

@inproceedings{DBLP:conf/nips/ZhangSZ24,
  author={Bozhou Zhang and Nan Song and Li Zhang},
  title={DeMo: Decoupling Motion Forecasting into Directional Intentions and Dynamic States},
  year={2024},
  cdate={1704067200000},
  url={http://papers.nips.cc/paper_files/paper/2024/hash/c0ff9e52e94ae331bc0f2d28be06a9ca-Abstract-Conference.html},
  booktitle={NeurIPS},
}

@article{huang2024trajectory,
  title={Trajectory Mamba: Efficient Attention-Mamba Forecasting Model Based on Selective SSM},
  author={Huang, Yizhou and Cheng, Yihua and Wang, Kezhi},
  journal={arXiv preprint arXiv:2503.10898},
  year={2024}
}

@inproceedings{
xu2025ppt,
title={{PPT}: Pretraining with Pseudo-Labeled Trajectories for Motion Forecasting},
author={Yihong Xu and Yuan Yin and Eloi Zablocki and Tuan-Hung Vu and Alexandre Boulch and Matthieu Cord},
booktitle={Workshop on Making Sense of Data in Robotics: Composition, Curation, and Interpretability at Scale at CoRL 2025},
year={2025},
url={https://openreview.net/forum?id=4SXdVmswuu}
}

@inproceedings{hu2023_uniad,
 title={Planning-oriented Autonomous Driving}, 
 author={Yihan Hu and Jiazhi Yang and Li Chen and Keyu Li and Chonghao Sima and Xizhou Zhu and Siqi Chai and Senyao Du and Tianwei Lin and Wenhai Wang and Lewei Lu and Xiaosong Jia and Qiang Liu and Jifeng Dai and Yu Qiao and Hongyang Li},
 booktitle={Proceedings of the IEEE/CVF Conference on Computer Vision and Pattern Recognition},
 year={2023},
}

@inproceedings{DBLP:conf/iccv/JiangCXLCZZ0HW23,
  author={Bo Jiang and Shaoyu Chen and Qing Xu and Bencheng Liao and Jiajie Chen and Helong Zhou and Qian Zhang and Wenyu Liu and Chang Huang and Xinggang Wang},
  title={VAD: Vectorized Scene Representation for Efficient Autonomous Driving},
  year={2023},
  cdate={1672531200000},
  pages={8306-8316},
  url={https://doi.org/10.1109/ICCV51070.2023.00766},
  booktitle={ICCV}
}

@InProceedings{Huang_2023_ICCV,
    author    = {Huang, Zhiyu and Liu, Haochen and Lv, Chen},
    title     = {GameFormer: Game-theoretic Modeling and Learning of Transformer-based Interactive Prediction and Planning for Autonomous Driving},
    booktitle = {Proceedings of the IEEE/CVF International Conference on Computer Vision (ICCV)},
    month     = {October},
    year      = {2023},
    pages     = {3903-3913}
}

@inproceedings{DBLP:conf/cvpr/LiYLLKL024,
  author={Zhiqi Li and Zhiding Yu and Shiyi Lan and Jiahan Li and Jan Kautz and Tong Lu and José M. Álvarez},
  title={Is Ego Status All You Need for Open-Loop End-to-End Autonomous Driving?},
  year={2024},
  cdate={1704067200000},
  pages={14864-14873},
  url={https://doi.org/10.1109/CVPR52733.2024.01408},
  booktitle={CVPR},
}

@inproceedings{
    jiang2026vadv,
    title={{VAD}v2: End-to-End Autonomous Driving via Probabilistic Planning},
    author={Bo Jiang and Shaoyu Chen and Hao Gao and Bencheng Liao and Qian Zhang and Wenyu Liu and Xinggang Wang},
    booktitle={The Fourteenth International Conference on Learning Representations},
    year={2026},
    url={https://openreview.net/forum?id=0a4dA6eUHN}
}

@article{zheng2024genad,
    title={GenAD: Generative End-to-End Autonomous Driving},
    author={Zheng, Wenzhao and Song, Ruiqi and Guo, Xianda and Zhang, Chenming and Chen, Long},
    journal={arXiv preprint arXiv: 2402.11502},
    year={2024}
}

@article{Sun2024SparseDriveEA,
  title={SparseDrive: End-to-End Autonomous Driving via Sparse Scene Representation},
  author={Wenchao Sun and Xuewu Lin and Yining Shi and Chuang Zhang and Haoran Wu and Sifa Zheng},
  journal={2025 IEEE International Conference on Robotics and Automation (ICRA)},
  year={2024},
  pages={8795-8801},
  url={https://api.semanticscholar.org/CorpusID:270123261}
}

@inproceedings{zhang2025bridging,
 title={Bridging Past and Future: End-to-End Autonomous Driving with Historical Prediction and Planning},
 author={Zhang, Bozhou and Song, Nan and Jin, Xin and Zhang, Li},
 booktitle={CVPR},
 year={2025},
}

@article{Song2025DontST,
  title={Don’t Shake the Wheel: Momentum-Aware Planning in End-to-End Autonomous Driving},
  author={Ziying Song and Caiyan Jia and Lin Liu and Hongyu Pan and Yongchang Zhang and Junming Wang and Xingyu Zhang and Shaoqing Xu and Lei Yang and Yadan Luo},
  journal={2025 IEEE/CVF Conference on Computer Vision and Pattern Recognition (CVPR)},
  year={2025},
  pages={22432-22441},
  url={https://api.semanticscholar.org/CorpusID:276782149}
}

@article{chen2024end,
  title={End-to-end autonomous driving: Challenges and frontiers},
  author={Chen, Li and Wu, Penghao and Chitta, Kashyap and Jaeger, Bernhard and Geiger, Andreas and Li, Hongyang},
  journal={IEEE Transactions on Pattern Analysis and Machine Intelligence},
  volume={46},
  number={12},
  pages={10164--10183},
  year={2024},
  publisher={IEEE}
}

@inproceedings{Li2022BEVFormerLB,
  title={BEVFormer: Learning Bird's-Eye-View Representation from Multi-Camera Images via Spatiotemporal Transformers},
  author={Zhiqi Li and Wenhai Wang and Hongyang Li and Enze Xie and Chonghao Sima and Tong Lu and Qiao Yu and Jifeng Dai},
  booktitle={European Conference on Computer Vision},
  year={2022},
  url={https://api.semanticscholar.org/CorpusID:247839336}
}

@article{shi2022motion,
  title={Motion transformer with global intention localization and local movement refinement},
  author={Shi, Shaoshuai and Jiang, Li and Dai, Dengxin and Schiele, Bernt},
  journal={Advances in Neural Information Processing Systems},
  volume={35},
  pages={6531--6543},
  year={2022}
}

@article{Zhou2023QueryCentricTP,
  title={Query-Centric Trajectory Prediction},
  author={Zikang Zhou and Jianping Wang and Yung-Hui Li and Yu-Kai Huang},
  journal={2023 IEEE/CVF Conference on Computer Vision and Pattern Recognition (CVPR)},
  year={2023},
  pages={17863-17873},
  url={https://api.semanticscholar.org/CorpusID:259359908}
}

@article{Cheng2024RethinkingIP,
  title={Rethinking Imitation-based Planners for Autonomous Driving},
  author={Jie Cheng and Yingbing Chen and Xiaodong Mei and Bowen Yang and Bo Li and Ming Liu},
  journal={2024 IEEE International Conference on Robotics and Automation (ICRA)},
  year={2024},
  pages={14123-14130},
  url={https://api.semanticscholar.org/CorpusID:271798811}
}

@inproceedings{dauner2023parting,
  title={Parting with misconceptions about learning-based vehicle motion planning},
  author={Dauner, Daniel and Hallgarten, Marcel and Geiger, Andreas and Chitta, Kashyap},
  booktitle={Conference on Robot Learning},
  pages={1268--1281},
  year={2023},
  organization={PMLR}
}

@article{Song2024RobustnessAware3O,
  title={Robustness-Aware 3D Object Detection in Autonomous Driving: A Review and Outlook},
  author={Ziying Song and Lin Liu and Feiyang Jia and Yadan Luo and Caiyan Jia and Guoxin Zhang and Lei Yang and Li Wang},
  journal={IEEE Transactions on Intelligent Transportation Systems},
  year={2024},
  volume={25},
  pages={15407-15436},
  url={https://api.semanticscholar.org/CorpusID:266977207}
}

@article{Wang2023MultiModal3O,
  title={Multi-Modal 3D Object Detection in Autonomous Driving: A Survey and Taxonomy},
  author={L. xilinx Wang and Xinyu Zhang and Ziying Song and Jiangfeng Bi and Guoxin Zhang and Haiyue Wei and Liyao Tang and Lei Yang and Jun Li and Caiyan Jia and Lijun Zhao},
  journal={IEEE Transactions on Intelligent Vehicles},
  year={2023},
  volume={8},
  pages={3781-3798},
  url={https://api.semanticscholar.org/CorpusID:260432447}
}

@article{Prakash2021MultiModalFT,
  title={Multi-Modal Fusion Transformer for End-to-End Autonomous Driving},
  author={Aditya Prakash and Kashyap Chitta and Andreas Geiger},
  journal={2021 IEEE/CVF Conference on Computer Vision and Pattern Recognition (CVPR)},
  year={2021},
  pages={7073-7083},
  url={https://api.semanticscholar.org/CorpusID:233148602}
}

@article{Lin2023Sparse4DVR,
  title={Sparse4D v2: Recurrent Temporal Fusion with Sparse Model},
  author={Xuewu Lin and Tianwei Lin and Zi-Hui Pei and Lichao Huang and Zhizhong Su},
  journal={ArXiv},
  year={2023},
  volume={abs/2305.14018},
  url={https://api.semanticscholar.org/CorpusID:258841133}
}

@article{Wang2023ExploringOT,
  title={Exploring Object-Centric Temporal Modeling for Efficient Multi-View 3D Object Detection},
  author={Shihao Wang and Yingfei Liu and Tiancai Wang and Ying Li and Xiangyu Zhang},
  journal={2023 IEEE/CVF International Conference on Computer Vision (ICCV)},
  year={2023},
  pages={3598-3608},
  url={https://api.semanticscholar.org/CorpusID:257636991}
}

@article{Zhang2024SparseADSQ,
  title={SparseAD: Sparse Query-Centric Paradigm for Efficient End-to-End Autonomous Driving},
  author={Diankun Zhang and Guoan Wang and Runwen Zhu and Jianbo Zhao and Xiwu Chen and Siyu Zhang and Jiahao Gong and Qibin Zhou and Wenyuan Zhang and Ningzi Wang and Feiyang Tan and Hangning Zhou and Ziyao Xu and Haotian Yao and Chi Zhang and Xiaojun Liu and Xiaoguang Di and Bin Li},
  journal={ArXiv},
  year={2024},
  volume={abs/2404.06892},
  url={https://api.semanticscholar.org/CorpusID:269033031}
}

@inproceedings{caesar2020nuscenes,
  title={nuscenes: A multimodal dataset for autonomous driving},
  author={Caesar, Holger and Bankiti, Varun and Lang, Alex H and Vora, Sourabh and Liong, Venice Erin and Xu, Qiang and Krishnan, Anush and Pan, Yu and Baldan, Giancarlo and Beijbom, Oscar},
  booktitle={Proceedings of the IEEE/CVF conference on computer vision and pattern recognition},
  pages={11621--11631},
  year={2020}
}

@article{jia2024bench2drive,
  title={Bench2drive: Towards multi-ability benchmarking of closed-loop end-to-end autonomous driving},
  author={Jia, Xiaosong and Yang, Zhenjie and Li, Qifeng and Zhang, Zhiyuan and Yan, Junchi},
  journal={Advances in Neural Information Processing Systems},
  volume={37},
  pages={819--844},
  year={2024}
}

@inproceedings{Hu2022STP3EV,
  title={ST-P3: End-to-end Vision-based Autonomous Driving via Spatial-Temporal Feature Learning},
  author={Shengchao Hu and Li Chen and Peng Wu and Hongyang Li and Junchi Yan and Dacheng Tao},
  booktitle={European Conference on Computer Vision},
  year={2022},
  url={https://api.semanticscholar.org/CorpusID:250607597}
}

@article{He2015DeepRL,
  title={Deep Residual Learning for Image Recognition},
  author={Kaiming He and X. Zhang and Shaoqing Ren and Jian Sun},
  journal={2016 IEEE Conference on Computer Vision and Pattern Recognition (CVPR)},
  year={2015},
  pages={770-778},
  url={https://api.semanticscholar.org/CorpusID:206594692}
}

@inproceedings{Loshchilov2017DecoupledWD,
  title={Decoupled Weight Decay Regularization},
  author={Ilya Loshchilov and Frank Hutter},
  booktitle={International Conference on Learning Representations},
  year={2017},
  url={https://api.semanticscholar.org/CorpusID:53592270}
}

@inproceedings{
    loshchilov2017sgdr,
    title={{SGDR}: Stochastic Gradient Descent with Warm Restarts},
    author={Ilya Loshchilov and Frank Hutter},
    booktitle={International Conference on Learning Representations},
    year={2017},
    url={https://openreview.net/forum?id=Skq89Scxx}
}
\end{document}